\def\year{2019}\relax
\documentclass[letterpaper]{article} 
\usepackage{aaai19}  
\usepackage{times}  
\usepackage{helvet}  
\usepackage{courier}  
\usepackage{url}  
\usepackage{graphicx}  
\usepackage{latexsym}
\usepackage{url}
\usepackage{graphicx}
\usepackage{epstopdf}
\usepackage{multirow}
\usepackage{amsfonts}
\usepackage{amsmath}
\usepackage{amssymb}
\usepackage{booktabs}
\usepackage[whole]{bxcjkjatype}
\usepackage[unicode]{hyperref}

\begin{document}
%
\title{Chinese Discourse Segmentation Using Bilingual Discourse Commonality}
\author{Jingfeng Yang \\
Institute of Computational Linguistics\\
Peking University\\
yjfllpyym@pku.edu.cn\\
\And Sujian Li\\
Institute of Computational Linguistics\\
Peking University\\
lisuajin@pku.edu.cn\\
}
\maketitle
\begin{abstract}
Discourse segmentation aims to segment Elementary Discourse Units (EDUs) and is a fundamental task in discourse analysis.
For Chinese, previous researches identify EDUs just through discriminating the functions of punctuations.
In this paper, we argue that Chinese EDUs may not end at the punctuation positions and should follow the definition of EDU in RST-DT.
With this definition, we conduct Chinese discourse segmentation with the help of English labeled data.
Using discourse commonality between English and Chinese,  we design an adversarial neural network framework to extract common language-independent features and language-specific features which are useful for discourse segmentation, when there is no or only a small scale of Chinese labeled data available.
Experiments on discourse segmentation demonstrate that our models can leverage common features from bilingual data,
and learn efficient Chinese-specific features from a small amount of Chinese labeled data, outperforming the baseline models.
\end{abstract}

\section{Introduction}


Discourse segmentation aims to segment Elementary Discourse Units (EDUs) which are defined as the minimal building blocks of constituting a text ~\cite{mann1988rhetorical}.
There exist many controversies about what constitutes an EDU.
A generally acceptable practice takes EDUs to be non-overlapping clauses~\cite{carlson2001discourse},
which has been verified a reasonable language unit and used successfully in some downstream applications such as automatic summarization~\cite{hirao2013single,li2016role,durrett2016learning}. 


For research on Chinese discourse, several discourse corpus have been constructed  ~\cite{zhou2012PDTB,li2014cdtb,zhang2014hit}.
\cite{zhou2012PDTB} built a predicate-argument style Chinese discourse treebank, following the annotation scheme of English Penn Discourse Treebank (PDTB) \cite{prasad2007penn}.
Motivated by Rhetorical Structure Theory (RST), \citeauthor{li2014cdtb}\shortcite{li2014cdtb} constructed a RST-like discourse treebank.
These work directly treated EDUs as Chinese clauses segmented by some punctuations, e.g. comma, semicolon and period.
At the same time, there are some work to research whether punctuations especially commas are boundaries of EDUs \cite{xue2011chinese,yang2012chinese,xu2013recognizing,li2015chinese}.

\begin{figure}
		
		\centering	
		%
		\includegraphics[width=1.0\columnwidth]{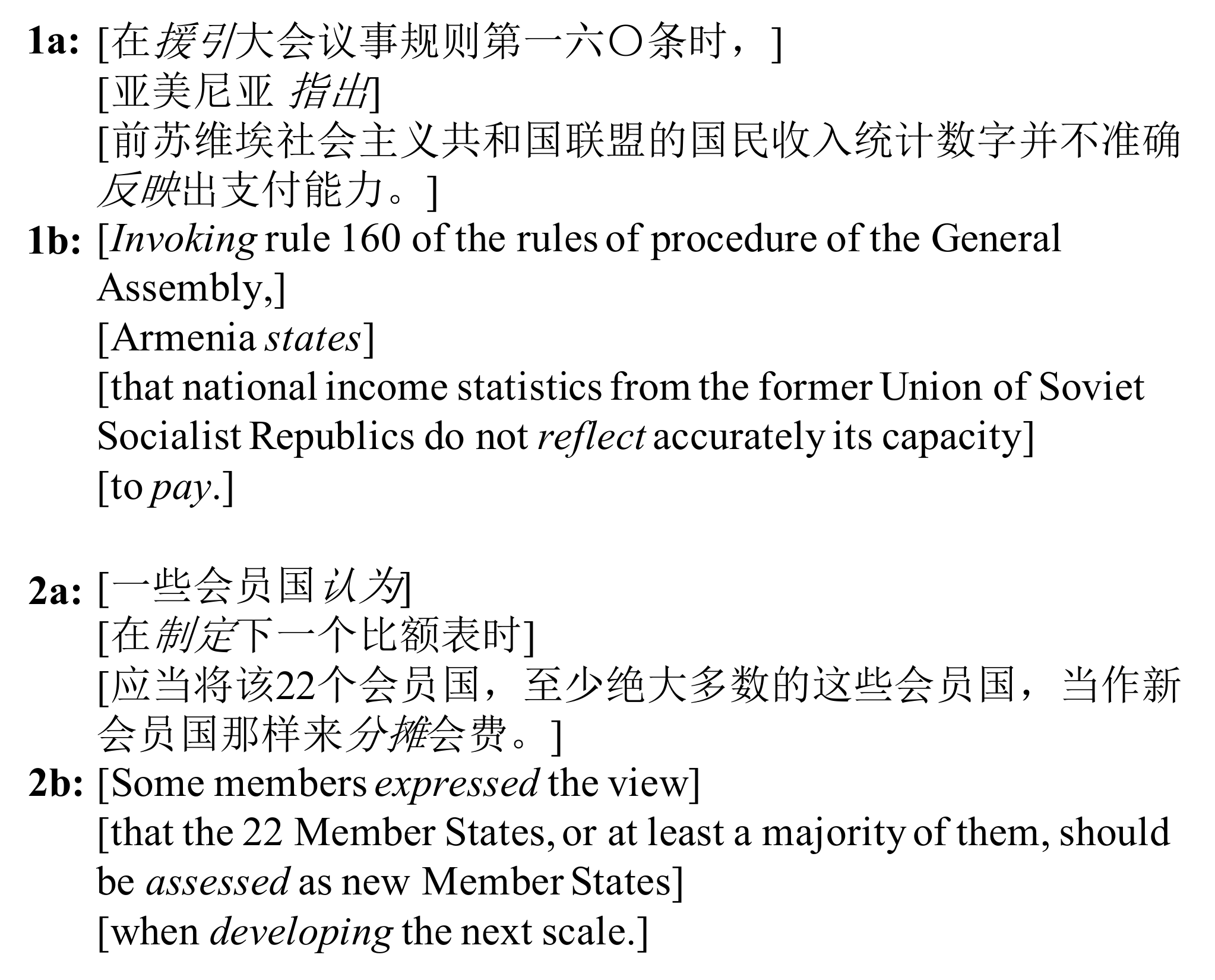}	
		\caption{Examples of Chinese and English discourse segmentation. Text included by [] means an EDU.
			The verb-like words are marked in italic font. }
\label{fig:intro-examp}		
\end{figure}

In our opinion, Chinese EDUs(or clauses)  may not always end at the punctuation positions.	
Fig.~\ref{fig:intro-examp} shows two Chinese examples of discourse segmentation (i.e., \textbf{1a} and \textbf{2a}). 
From the examples we can see that, besides punctuations, some intra-sentential words may be the boundaries of EDUs,
e.g.,  ``亚美尼亚指出 (Armenia state)'' in \textbf{1a} and ``在制定下一个比额表时(when developing the next scale)'' in \textbf{2a}.
	
As we know, the discourse guidelines defined by \cite{carlson2001discourse}  compile a series of rules for segmenting English EDUs, which can also apply on Chinese text.
However, compared to English, it is relatively difficult to identify Chinese EDUs under this guideline,
because Chinese phrase structure is syntactically similar to sentence structure, and relative pronouns and subordinating conjunctions are rarely used to introduce clauses~\cite{tsao1990}.
Thus, in this paper we argue to adopt the EDU segmentation guideline proposed by by \cite{carlson2001discourse} and aim to conduct Chinese discourse segmentation  with a more appropiate way.



For our work, the main problem is the absence of labeled Chinese data which conforms to this segmentation guideline.
It is fortunate that  there exist some  discourse commonalities between Chinese and English as we observe.
Figure~\ref{fig:intro-examp} gives two English discourse segmentation examples (i.e., \textbf{1b} and \textbf{2b}) which are translations of \textbf{1a} and \textbf{2a} respectively.
We can see that each EDU in either language contains a verb-like word and other similar syntactic and lexical features in judging the EDU boundary.
In such cases, we come up with the idea that Chinese EDU segmentation can be conducted with the help of the discourse commonality summarized from substantial English discourse data such as RST Discourse Treebank (RST-DT) ~\cite{carlson2001discourse}.  

As for cross-lingual discourse segmentation, ~\citeauthor{braud2017cross}\shortcite{braud2017cross} designed a multi-task learning framework for five languages (i.e., English, Spanish, German, Dutch and Brazilian Portuguese) which belong to the Indo-European  family and are very similar.
Their model is relatively simple and can not well extract common features for discourse segmentation sharing in much different languages such as Chinese and English, since Chinese belongs to Sino-Tibetan language faminly and Engligh.
To transfer English segmentation information to Chinese, inspired by the work of ~\cite{ganin2015unsupervised}, we design the adversarial neural network framework which can leverage both language-specific  features and shared language-independent features for discourse segmentation.
We propose to use English labeled data to train a discourse segmenter and simultaneously transfer the learned discourse segmentation features to  Chinese.
A common Bidirectional Long-short term memory (BiLSTM) network is designed with language-adversarial training to exploit the language-independent knowledge which can be transferred from English to Chinese. At the same time, two private BiLSTMs are used to extract language-specific knowledge.

To the best of our knowledge, we are the first to propose that Chinese EDU segmentation can not be limited to only judging the functions of punctuations and contribute to providing a small scale of Chinese labeled corpus.
We also contribute to using an adversarial neural network framework to conduct Chinese discourse segmentation with the help of English labeled data.
Experiments show that our models can leverage common features from English data,
and learn efficient Chinese-specific features from a small amount of Chinese labeled data, outperforming the baseline models.

\section{Chinese Elementary Discourse Units}
\label{sec:resourse}

In this section, we briefly introduce the definition of Chinese EDUs in our work.
Generally, the basic principle is to treat an EDU as a clause. 
However, since a discourse unit is a semantic concept while a clause is syntactically defined, researchers further made some refinements on EDU segmentation.
We mainly followed the guidelines defined by ~\citeauthor{carlson2001discourse}\shortcite{carlson2001discourse} and manually labeled a small corpus according to the characteristics of Chinese.
Next we list some criteria of segmenting EDUs as follows.
\begin{itemize}
\setlength{\itemsep}{0pt}	
\item[1.] \textbf{Subjective clauses}, objective clauses of non-attributive  verbs and verb complement clauses are \textbf{not} segmented as EDUs.
\item[2.] \textbf{A prepositional clause} is an EDU.
\item[3.] Complements of \textbf{attribution verbs}, including both speech acts and other cognitive acts, are treated as EDUs.
\item[4.] \textbf{Coordinated sentences and clauses} are broken into separate EDUs.
\item[5.] \textbf{Coordinated verb phrases} are \textbf{not} separated into EDUs.
\item[6.] \textbf{Temporal expressions} are marked as separate EDUs.
\item[7.] Correlative subordinators consist of two separate EDUs, provided that \textbf{the subordinate clause} contains a verbal element.
\item[8.] \textbf{Strong discourse cues} can start a new EDU no matter they are followed by a clause or a phrase.
\item[9.] There are some embedded discourse units where one EDU is seperated by a \textbf{subordinate clause}.
\end{itemize}

\begin{figure}[!htb]
	\includegraphics[width=1.0\columnwidth]{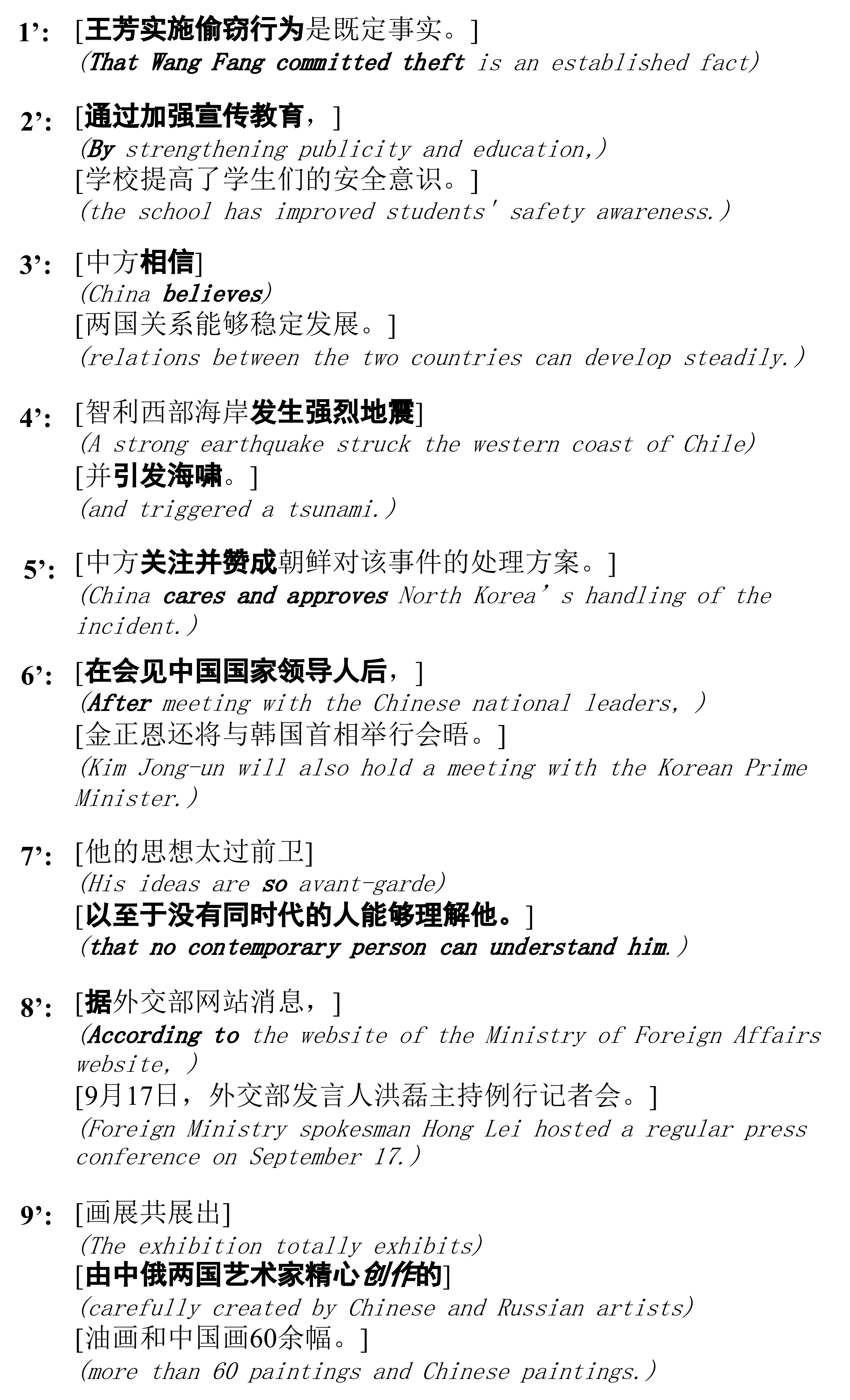}
	\setlength{\abovecaptionskip}{0cm}
	\setlength{\belowcaptionskip}{0cm}	
	\caption{Examples 1'$\sim$9' of EDU segmentation correspond to the criteria 1$\sim$9.
	}
	\label{fig:rule-example1}
\end{figure}

We also show some EDU segmentation examples in Fig. \ref{fig:rule-example1}, which correspond to the above segmentation criteria in bold face.
Due to space limitation, we do not explain the segmentation criteria in detail.
Following the  EDU segmentation guideline, we  manually segment 782 sentences from \textit{People's Daily} for training, validation and test.

\section{Method}
\label{sec:method}
\subsection{Model Architecture}

\begin{figure*}	
\centering
\includegraphics[width=1.6\columnwidth]{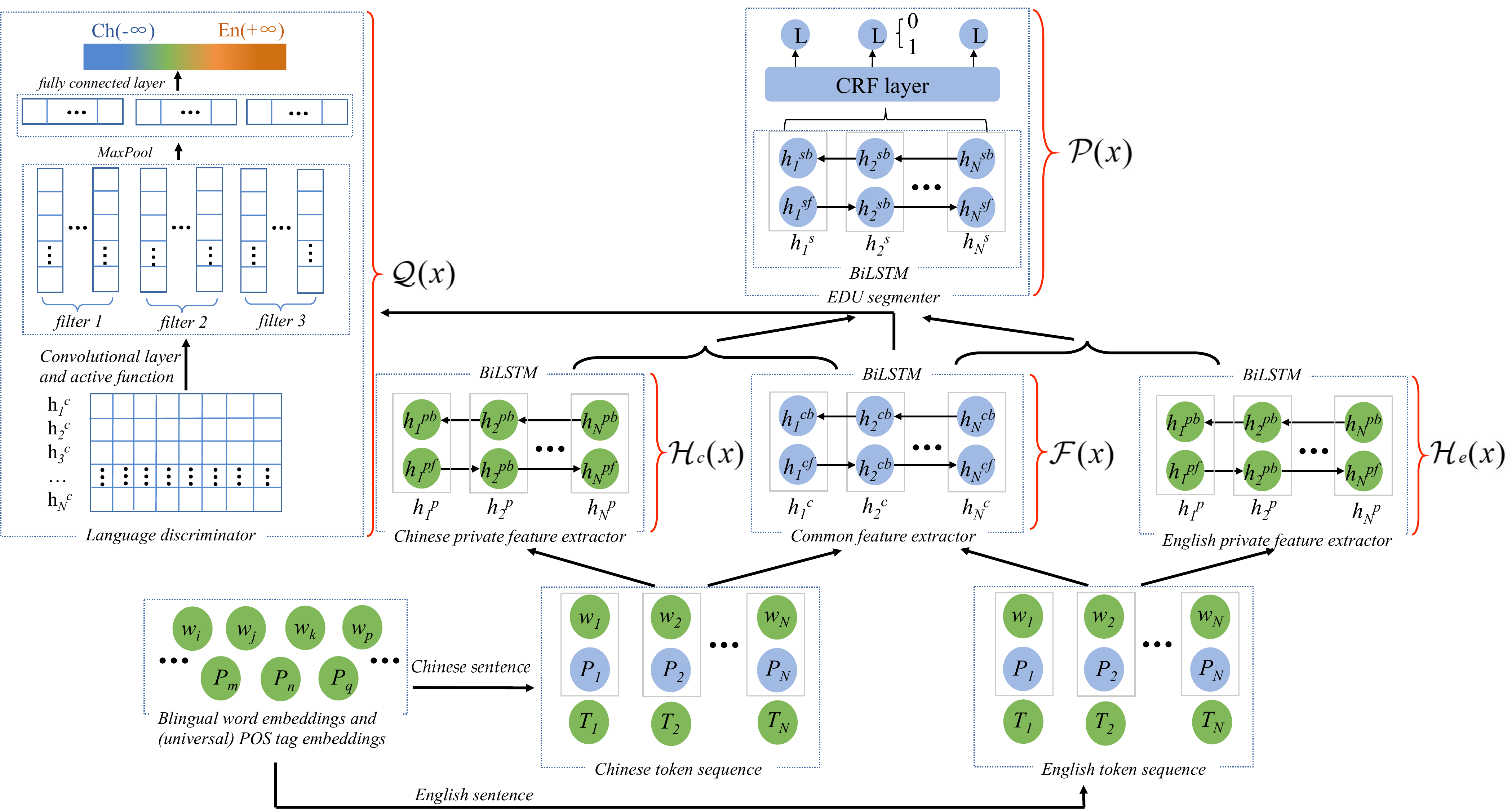}
\caption{Model architecture}
\label{fig:model}
\end{figure*}

To conduct Chinese discourse segmentation by exploiting English labeled data, we design our model as in Fig. \ref{fig:model}.
The whole architecture is mainly composed of four modules:
\textit{a common feature extractor} ($\mathcal{F}$) that extracts the shared language-independent features for discourse segmentation,
\textit{private feature extractors} ($\mathcal{H}_c$ and $\mathcal{H}_e$) that learns language-specific features respectively for Chinese and English,
an \textit{EDU segmenter} ($\mathcal{P}$) that conducts discourse segmentation for a sequence,
and a \textit{language discriminator} ($\mathcal{Q}$) that judges whether a sequence is in Chinese or English language.

To be specific,  given a sentence $x$ composed of $N$ words, we look up the pre-trained bilingual word embeddings $W_1, W_2, ..., W_N$ and Universal POS tag embeddings $P_1, P_2, ..., P_N$, and use their concatenation ($X=T_1, T_2, ..., T_N$) as the input of three feature extrators (i.e.,$\mathcal{F}$, $\mathcal{H}_c$ and $\mathcal{H}_e$ ), each of which is modeled using the bidirection LSTMs (BiLSTM) models.
Knowing the used language of $x$, $X$ is inputted into the common feature extractor and the corresponding language-specific feature extractor.
The common feature extractor maps $X$ into the hidden states $\mathcal{F}(x) = h_1^c, h_2^c, ..., h_N^c$ which contains both forward and backward information learned by  LSTMs.
Similarly, the private feature extractor converts $X$ into the hidden states $\mathcal{H}_c(x) = h_1^{p_c}, h_2^{p_c}, ..., h_N^{p_c}$ if $x$ is in Chinese, $\mathcal{H}_e(x) = h_1^{p_e}, h_2^{p_e}, ..., h_N^{p_e}$ if $x$ is in English.
Next, with the hidden states as input, we design two tasks: one task is discourse segmentation and the other is a language-adversarial task.

$\bullet$\textbf{Discourse segmentation task.}
Discourse segmentation is regarded as a sequence labeling problem with two labels (i.e., 0 and 1) indicating whether to segment the current word.
With the concatenation of $\mathcal{F}(x)$ with $\mathcal{H}(x)$ (i.e., $\mathcal{H}_c(x)$ or  $\mathcal{H}_e(x)$),
a BiLSTM layer is applied to obtain the abstract hidden states 
$h_1^s, h_2^s, ..., h_N^s$
for the following CRF layer which outputs the final labeling results.
Suppose $\mathcal{T}$ is the transition matrix of a linear-CRF and $e$, $b$ are score vectors that capture the cost of beginning and ending with a given label. Then, the global score of a given label sequence $y$ is represent by
\begin{equation}
\begin{split}
\mathcal{P}(\mathcal{F}(x),\mathcal{H}(x),y){\equiv}&b[y_1]+\sum_{i=1}^{m} [ h_i^s \rightarrow y_i]+\\
&\sum_{i=1}^{m-1}\mathcal{T}[y_i,y_{i+1}]+e[y_m]
\end{split}
\end{equation}
Where $[ h_i^s \rightarrow y_i]$ denotes the cost of labeling $y_i$ with the hidden representation of $h_i^s$.

Then, we choose the best label sequence based on the global score.
Let the parameters of the discourse segmenter be $\Theta_p$.
Given $\mathcal{F}$ and $\mathcal{H}$, we use cross-entropy loss as the objective function:
\begin{equation}
\begin{split}
{J_p}{\equiv}\min\limits_{\Theta_p}\mathbb{E}_{(x,y)}-\log\frac{e^{\mathcal{P}(\mathcal{F}(x),\mathcal{H}(x),y)}} {\sum_{y} e^{\mathcal{P}(\mathcal{F}(x),\mathcal{H}(x),y)}}
\end{split}
\end{equation}

$\bullet$\textbf{Language-adversarial task.}
The language-adversarial task aims to train the language-independent features which make the language discriminator difficult to distinguish what language is used.
The idea is inspired by domain adversarial training and Generative Adversarial Network (GAN), which has been applied on sentiment classification and POS tagging ~\cite{chen2016adversarial,kim2017cross}.
The language discriminator is composed of convolution neural networks (CNNs) and a fully-connected layer with sigmoid activation function, with reference to the text classification model~\cite{kim2014convolutional}.

First, the common language-independent features $\mathcal{F}(x) = h_1^c, h_2^c, ..., h_N^c$ are inputted to three convolution filters  whose  activation function is Leaky ReLU ~\cite{maas2013rectifier} and window sizes are set to 3, 4 and 5 respectively. Then, the max-over-time pooling (MaxPool) operation is applied to obtain three fixed-length vectors which are concatenated as a vector.
Next, the fully connected layer takes the concatenated vector as input and outputs a scalar value. The higher the value is, the more probable the language is in English.

When training the language discriminator, we expect the scores for English instances to be higher and the scores for Chinese to be lower. 
Here, we define the distribution of the common language-independent features $\mathcal{F}$ for $English$ and $Chinese$ instances as:
\newenvironment{shrinkeq}[1]
{ \bgroup
\addtolength\abovedisplayshortskip{#1}
\addtolength\abovedisplayskip{#1}
\addtolength\belowdisplayshortskip{#1}
\addtolength\belowdisplayskip{#1}}
{\egroup\ignorespacesafterend}

\begin{shrinkeq}{-3ex}
\begin{equation}\nonumber
\begin{split}
P_\mathcal{F}^{Ch}{\triangleq}P(\mathcal{F}(x){\vert}x{\in}Chinese)\\
P_\mathcal{F}^{En}{\triangleq}P(\mathcal{F}(x){\vert}x{\in}English)
\end{split}
\end{equation}
\end{shrinkeq}

~\\
We train $\mathcal{F}(x)$ to make these two distributions as close as possible so that  $\mathcal{F}(x)$ is independent of language.
Following ~\cite{chen2016adversarial} and ~\cite{arjovsky2017wasserstein}, we minimize the  Wasserstein distance between $P_\mathcal{F}^{Ch}$ and $P_\mathcal{F}^{En}$ since the Wasserstein distance is relatively stable  for hyperparameter selection.
We notate the language discriminator as the function $\mathcal{Q}$,
according to the Kantorovich-Rubinstein duality ~\cite{villani2008optimal} we get the  Wasserstein distance $W(P_\mathcal{F}^{En},P_\mathcal{F}^{Ch})$:
\begin{equation}
\begin{split}
&W(P_\mathcal{F}^{En},P_\mathcal{F}^{Ch})=\sup_{{\mathcal{Q}}}\\
&\mathbb{E}_{\mathcal{F}(x){\sim}P_\mathcal{F}^{En}}[\mathcal{Q}(\mathcal{F}(x))]-\mathbb{E}_{\mathcal{F}(x'){\sim}P_\mathcal{F}^{Ch}}[\mathcal{Q}(\mathcal{F}(x'))]
\end{split}
\end{equation}


Let the discriminator $\mathcal{Q}$ be parameterized by $\Theta_q$, and the objective function $J_q$ given $\Theta_q$ is:
\begin{equation}
\begin{split}
&{J_q}{\equiv}\max \limits_{\Theta_q}\\
&\mathbb{E}_{\mathcal{F}(x){\sim}P_\mathcal{F}^{Eng}} [\mathcal{Q}(\mathcal{F}(x))]-\mathbb{E}_{\mathcal{F}(x'){\sim}P_\mathcal{F}^{Chn}}[\mathcal{Q}(\mathcal{F}(x'))]
\end{split}
\end{equation}

Finally, when training the  parameters of the common feature extractor $\mathcal{F}(x)$, we minimize the objective function ${J_q}$  given that the discriminator has been trained well. This means that the common features between English and Chinese are language-independent and the discriminator have no capacity to distinguish the two languages.
At the same time, we need to consider the discourse segmentation task with $\mathcal{F}(x)$  and $\mathcal{H}(x)$ as input.
Let the feature extractors be parameterized as ($\Theta_f,\Theta_h$), and we train the features with the objective function:
\begin{equation}
\begin{split}
{J_{f,h}}{\equiv}\min\limits_{\Theta_f,\Theta_h}{J_p}({\Theta_f,\Theta_h})+\lambda{J_q}({\Theta_f})
\end{split}
\end{equation}

\subsection{Model Training}
\label{subsec:training}
To train our model, we use English labeled data, Chinese unlabeled data and zero (or a little) Chinese labeled data. The English labeled corpus is important for learning language-independent features. According to the Chinese resource we use, two ways of training are designed.

$\bullet$\textbf{With zero Chinese labeled  data (ZL)}. We only use  English labeled corpus and  Chinese unlabeled corpus to train the language discriminator ($\Theta_q$), common feature extractor ($\Theta_f$) and EDU segmenter ($\Theta_p$). 
Since no labeled Chinese data is used, the private Chinese-specific feature extractor can not be trained.
Only the common language-independent features are fed into the segmenter for Chinese discourse segmentation.

$\bullet$\textbf{Only with a little Chinese labeled data (LL)}.
We use English and Chinese labeled data to train our model and do not use Chinese unlabeled data. To balance the scale of Chinese data with that of English data, we keep duplicating  the small amount of Chinese labeled data until it has a similar scale with the English labeled data.
The duplicated Chinese labeled data and English labeled data are used to train our model until the Chinese validation data reaches the highest performance.

\section{Experiments}
\subsection{Resources}

\noindent\textbf{Chinese data.}
In this work of Chinese discourse segmentation, there are totally 782 manually labeled Chinese sentences, which were introduced in section \ref{sec:resourse}.  Among these data, we choose 182 sentences to compose of a  test set and 200 sentences a validation set , and use the rest 400 sentences as training data when needed. Besides, another 9236 sentences from  \textit{People's Daily} serves as unlabeled Chinese training data.

\noindent\textbf{English data.}
We use the RST-DT corpus composed of 385 discourse-segmented articles from Wall Street Journal (WSJ) and 500 segmented articles from ~\cite{yang2018scitb}.
Finally, we totally get 9636 segmented English sentences.
\noindent\textbf{Bilingual word embeddings (BWE).}
To obtain cross-lingual word embeddings which share the same space between Chinese and words, we use BilBOWA ~\cite{gouws2015bilbowa} to obtain the 200-dimensional bilingual word embeddings. To train BilBOWA, we use 570,499 Chinese sentences from \textit{People's Daily}, 700,000 English sentences from CNN/DailyMail ~\cite{hermann2015teaching}, and a parallel corpus composed of 50,000 pairs of aligned Chinese and English sentences.

%

\noindent\textbf{Universal POS tags.}
In order to assure that Chinese and English adopt the same POS tagset, we use Universal POS tagset ~\cite{nivre2016universal}.
For English text, we use pre-trained UDPipe model ~\cite{straka2016udpipe} to postag.
For Chinese text, we use Stanford CoreNLP toolkit ~\cite{manning2014stanford} to postag with the UPenn tagset and then convert the tag to Universal POS tag with a conversion map\footnote{Refer to \url{http://universaldependencies.org/tagset-conversion/zh-conll-uposf.html} and make some modifications.}, 
because UDPipe lacks Chinese training data and performs poorly.

\subsection{Setup}


When training our model, we train the feature extractors ($\mathcal{F},\mathcal{H}$)  and the discourse segmenter ($\mathcal{P}$) together, and train the language discriminator ($\mathcal{Q}$) separately. $\mathcal{F},\mathcal{H}$ and $\mathcal{P}$ are trained $k$ iterations per $\mathcal{Q}$ iteration. Following the ANDN network ~\cite{chen2016adversarial}, we set $k$ to be the value of 5.
For the parameter $\lambda$ which balances the influence of  $\mathcal{P}$ and $\mathcal{Q}$, we experiment the values of  $\{$0.2, 0,1, 0.05$\}$ on the validation data and finally set it to be 0.1. All the models are optimized using RMSProp with learning rate of 0.001 and decaying rate of 0.9. The reason why we do not use Adam is that momentum-based optimizers are unreliable when training weight-clipped WGAN ~\cite{arjovsky2017towards}. When training $\mathcal{Q}$, the threshold of weight-clipping is set as 0.01. The word embeddings dimension is set 200. The POS tag embeddings are initialized randomly and their dimension is set 50. The dimensions of the hidden states in the BiLSTMs are set the same as the dimensions of their corresponding input vectors. Besides, the outputs of all BiLSTMs are regularized with dropout rate of 0.5 ~\cite{pham2014dropout}. The size of minibatch is set 20.
To verify the effectiveness of our model, we design two sets of experiments. In the first set of experiments, we compare our models with several baselines and exhibit the performance improvement brought by the discourse commonality between Chinese and English and the adversarial language training model. In the second set of experiments, we analyze the factors such as size of the labeled Chinese data  which may influence the performance of our model. In the experiments, we use Precision, Recall and F-measure to evaluate the performance and F-measure is the main metric of measuring the quality of a model.

\subsection{Results}
\label{subsec:result}
Since there is no available tools which follow the RST-DT criteria to segment Chinese EDUs, here we design several baselines for comparison.
First, considering when there is no Chinese labeled training data, we use three baselines to contrast with our model of \textit{ZL}. The first baseline is named \textit{S-baseline} in which we directly see each sentence as an EDU. The second baseline is named \textit{P-baseline} which segments the sentences whenever it meets a punctuation except quotation marks (i.e., “ ”) and slight-pause mark (i.e., 、). The third is a baseline named \textit{NoAdvers-Z} which is identical to our model except the adversarial training part is removed. In another view, NoAdvers-Z can be seen as a segmenter which is composed of two-layer LSTM and one layer of CRF. We use English labeled data to train NoAdvers-Z and test it on Chinese text which is similar to the method of ~\cite{braud2017cross}. When we use the small amount of Chinese labeled  data,  we design a baseline \textit{NoAdvers-L} to contrast with our model \textit{LL}. The only difference between NoAdvers-L and NoAdvers-Z is that we train NoAdvers-L using the duplicated Chinese data as training set.

\begin{table}[]
\centering

\begin{tabular}{@{}cccc@{}}
\toprule
Method & F-measure & Precision & Recall    \\ \midrule
S-baseline                                                 & 48.46\%    & 100.00\%    & 48.46\%      \\
P-baseline                                                 & 79.41\%    & 89.76\%    & 71.20\%      \\
NoAdvers-Z                                                & 71.13\%    & 77.50 \%    & 65.72 \%      \\
ZL                                                        & 82.35\%    & 87.33\%    & 77.92\%     \\
\midrule
NoAdvers-L                                                & 86.54\%    & 91.19\%    & 82.33\%      \\
LL                                                        & \textbf{87.70\%}    & 93.59\%    & 82.51\%    \\
\bottomrule
\end{tabular}
\caption{Results Comparison}
\label{tab:results}
\end{table}

Table ~\ref{tab:results} shows the results of our models with the baselines. From the table, we can see \textit{S-baseline} achieves the worst performance, though its precision is 100 percent because each sentence boundary must be an EDU boundary. The recall of 48.46\% means that only 48.46\% of all EDU boundaries belong to the sentence boundaries. \textit{P-baseline} achieves a high F-measure of 79.41\%, though it is simple. According to the performance of \textit{P-baseline}, we know that 71.2\% of EDU boundaries belong to the punctuations and most of the punctuations (89.76\%) prefer to be EDU boundaries. Thus, the most challenging in Chinese discourse segmentation is to recall the rest 28.8\% EDU boundaries and pick out those punctuations which do not serve as EDU boundaries. We can also see that \textit{NoAdvers-Z} performs worse than \textit{P-baseline} with regard to the three metrics of Precision, Recall and F-measure. This means that we can not directly use the English training data to train a Chinese discourse segmenter even if bilingual embeddings of two languages share the same space. \textit{ZL} significantly outperforms the other three baselines when no labeled Chinese data is available and achieves the F-measure of 82.35\% by leveraging the common feature extractor adversarial network.
\begin{figure}	
\centering
\includegraphics[width=1.0\columnwidth]{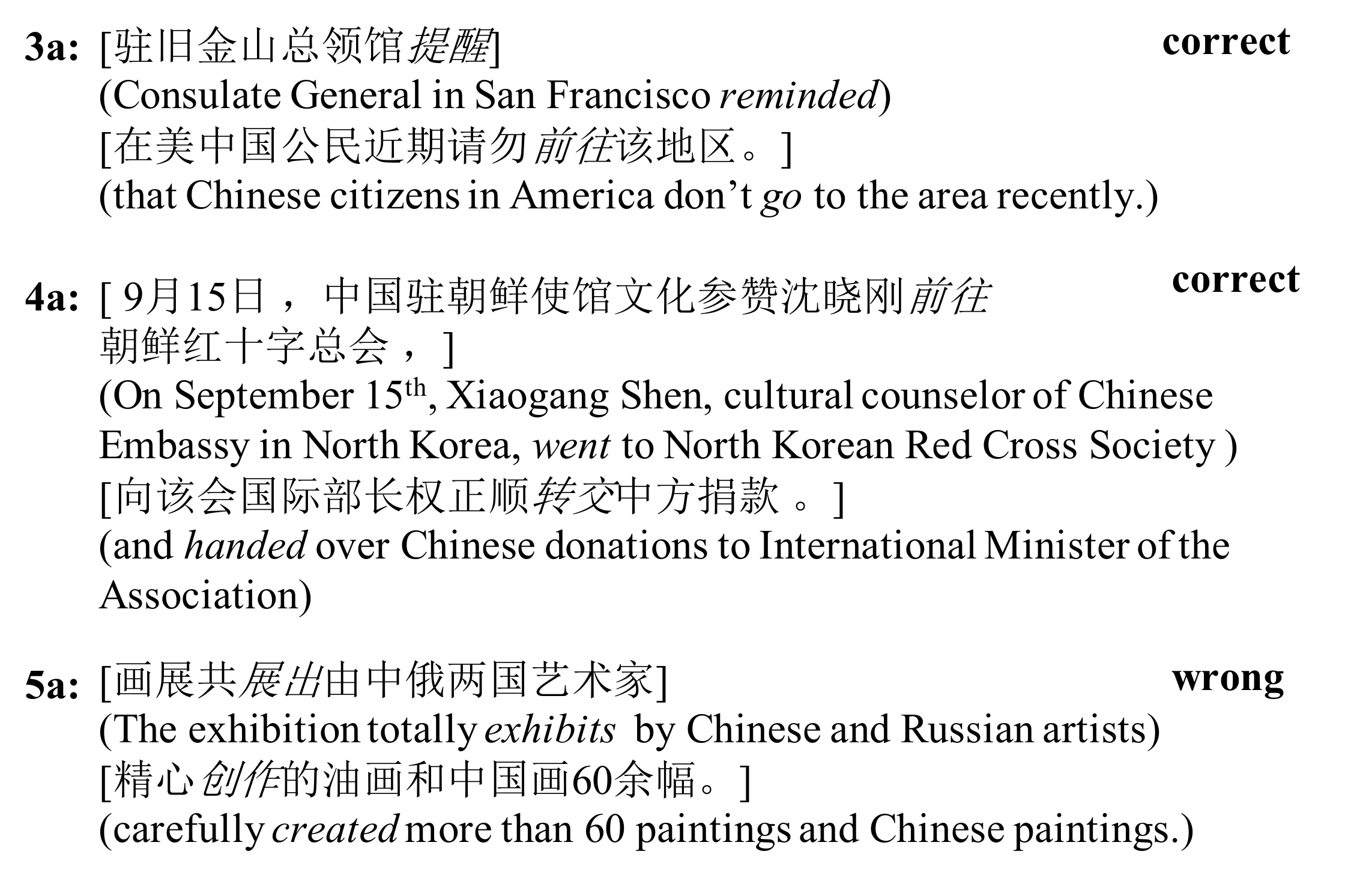}
\caption{Examples of EDU segmentation by \textbf{ZL} model. For each EDU, we give its literal English translation in parenthesis. \textbf{3a}, \textbf{4a} and \textbf{5a} are system results. \textbf{3a} and \textbf{4a} are correct while \textbf{5a} is wrong.   }
\label{fig:WZL}
\end{figure}
Compared to \textit{P-baseline}, \textit{ZL} can identify more EDU boundaries because the shared language-independent features learned from English labeled data are more appropriate to recognizing Chinese EDUs.  This can be verified from several segmentation examples which are shown in Figure~\ref{fig:WZL}.
We can see that Example (\textbf{3a}) can identify the attribution verb ``提醒 (remind)'', which indicates an EDU boundary following a subordinate clause. This means that ``提醒 (remind)'' is also an attribution verb in the English training data and the Chinese segmenter learns the feature.
Example (\textbf{4a}) correctly judges the functions of two commas where the content before the first comma is a temporal expression
and the second comma is a real EDU boundary, because our model may learn the language-independent features that a temporal clause can not be a separate EDU and verb-like words should be contained in an EDU. It is obvious that \textit{ZL} has its limitations.
Example (\textbf{5a}) is wrongly segmented and each segmented EDU does not have a complete meaning, though the model learns to contain a verb-like word in each EDU. This is because Chinese is a language of parataxis and use the passive voice with adding a function word ( e.g., ``由...创作(be created)''   in this example), which is different from English. It is difficult for \textit{ZL} to learn the Chinese-specific features without any Chinese labeled data.

\begin{figure}	
\centering
\includegraphics[width=\columnwidth]{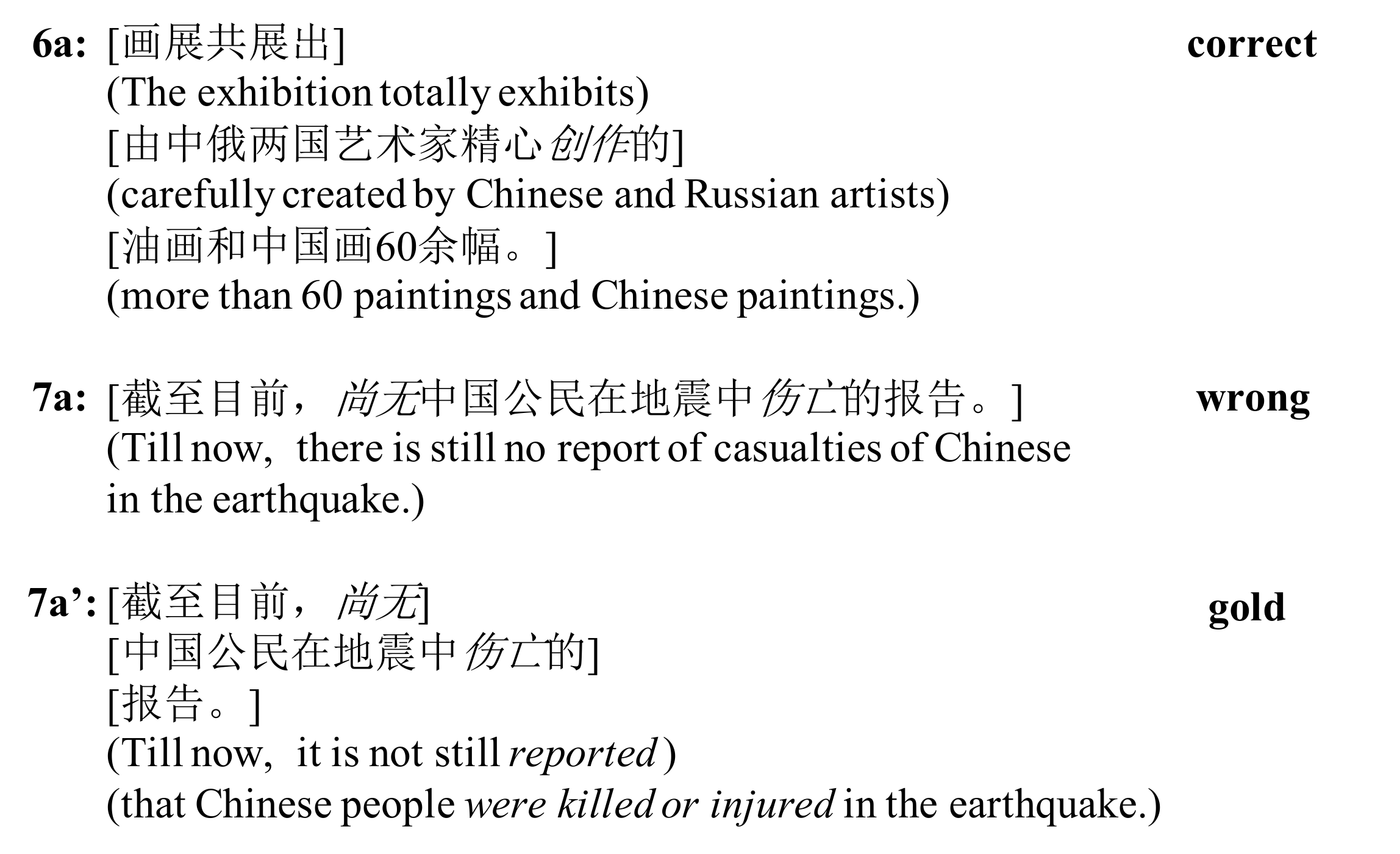}
\caption{Examples of EDU segmentation by \textbf{LL} model. For each EDU, we give its literal English translation in parenthesis. \textbf{6a} and \textbf{7a} are system results. \textbf{6a} is correct while \textbf{7a} is wrong. \textbf{7a'} is the gold segmentation result of \textbf{7a}. }
\label{fig:WLU}
\end{figure}
Table \ref{tab:results} also shows the performance of \textit{NoAdvers-L} and \textit{LL}, which use the 400 Chinese labeled sentences to supervise the training. From the table, we can see that \textit{NoAdvers-L} and \textit{LL} both outperform over the methods using zero Chinese labeled data. With Chinese and English labeled data together, \textit{LL} can effectively train the Chinese-specific features as well as the common language-independent features which are useful to labeling the EDU boundaries. At the same time, \textit{NoAdvers-L} achieves satisfactory results and again verifies that labeled data can provide effective supervision. \textit{LL} outperforms \textit{NoAdvers-L}, since the common features learned through the adversarial language discrimination and cross-lingual resources remedy training insufficiency of Chinese-specific features caused by the small Chinese labeled data.
Figure ~\ref{fig:WLU}  shows two segmentation examples by the \textit{LL} model. Here \textit{LL} correctly segments Example (\textbf{6a}) which has been wrongly segmented by \textit{ZL} (i.e., \textbf{5a}). From this example, we can see that the Chinese labeled data can train more Chinese-specific features such as the embedded attributive clause. Example (\textbf{7a})is still wrongly segmented by \textit{LL}. Here ``报告 (report)'' and ``伤亡 (casualty)''  are seen as noun words. In fact, when our annotators read this sentence,  ``伤亡'' is always understood as a verb phrase ``be killed or injured''. This is because verbs and verb phrases in Chinese can be nominalized without any overt marker for it. Without a large amount of Chinese labeled data, it is difficult to learn such ambiguous knowledge which is crucial to discourse segmentation.


\subsection{Impact of Resources}
\begin{figure}	
\centering
\includegraphics[width=0.8\columnwidth]{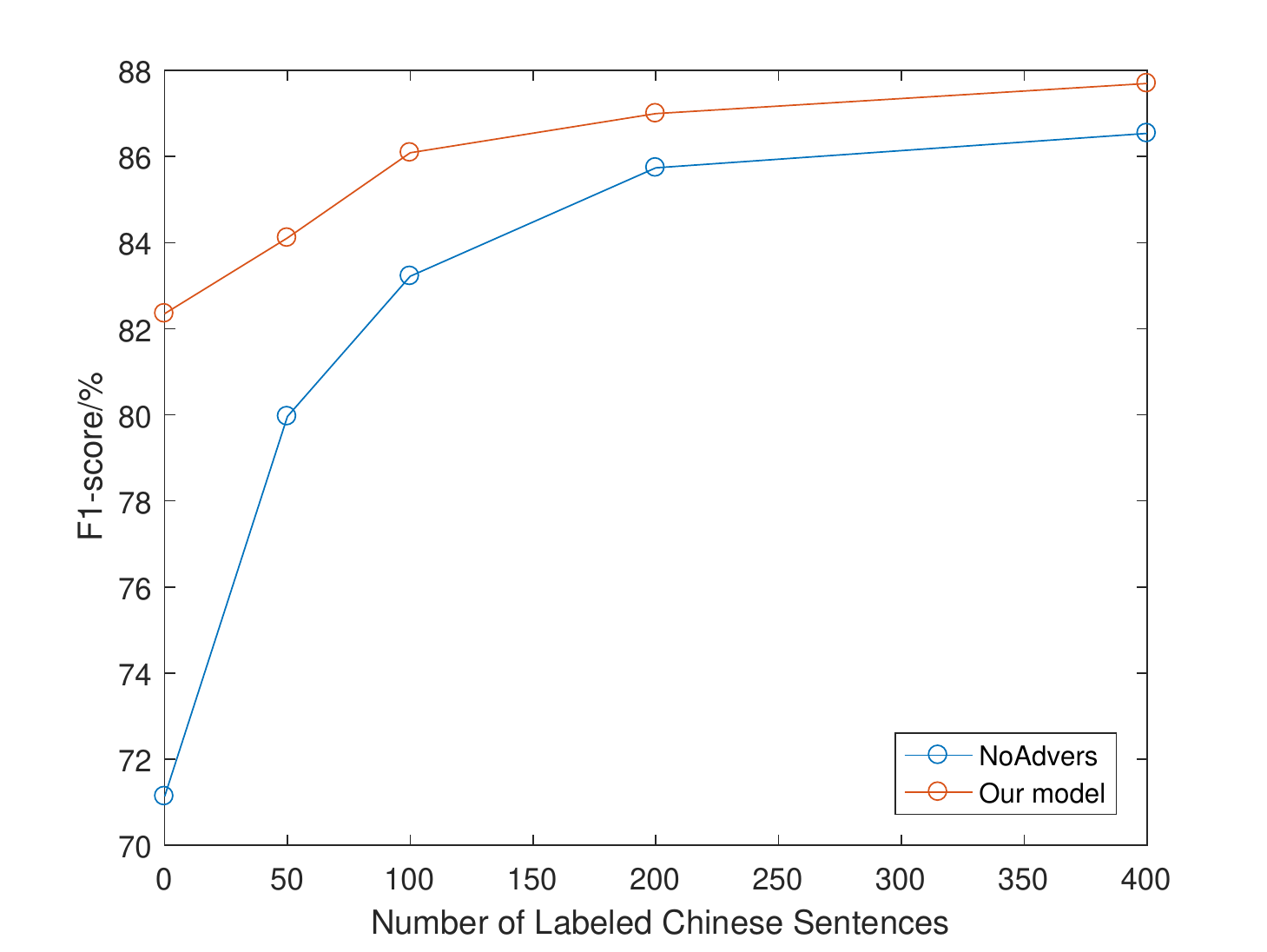}
\caption{Impact of labeled Chinese resource scale}
\label{fig:size}
\end{figure}
In this subsection, we mainly investigate the influence of size of Chinese labeled data and Universal POS tagset.
To figure out to what extend our model can remedy the lack of labeled Chinese data, we explore the performance change with regard to the size of these data and compare our models with the \textit{NoAdvers} model. Figure~\ref{fig:size} displays the performance change. With zero Chinese labeled data, there is a large performance gap between our \textit{ZL} model and the \textit{NoAdvers-Z} model, because the common features learned from the adversarial mechanism contribute much to Chinese discourse segmentation in \textit{ZL}.
With Chinese labeled data increasing, the performance gap between our \textit{LL} model and the \textit{NoAdvers-L} model first becomes decreased and then keeps stable when the size of Chinese labeled data reaches 200 sentences. The performance tendency means that Chinese labeled data when available is more effective for training the Chinese discourse segmenter than English data. The more Chinese labeled data a model has, the higher performance it can boost. From the performance gaps between our model and the \textit{NoAdvers} model, we can say that the common language-independent features learned from cross-lingual discourse commonality also contributes to Chinese discourse segmentation. Since we do not have a large amount of Chinese labeled data, we are not sure whether the two performance lines will intersect and whether the common features learned from English data will not contribute to Chinese discourse segmentation when Chinese labeled data is large enough. At least, with the small-scale Chinese training data in hand, the common language-independent features learned from the adversarial discrimination are still important to the performance of a Chinese discourse segmenter.

\begin{table}[]
\centering
\begin{tabular}{@{}ccc@{}}
\toprule
POS tagset          & \textbf{ZL}     &   \textbf{LL} \\ \midrule
Universal  & 82.35\%                                            & 87.70\%                                              \\
Different  & 79.40\%                                            & 86.37\%                                              \\ \bottomrule
\end{tabular}
\caption{Impact of POS tagset}
\label{tab:tagset}
\end{table}

To show the necessity of using Universal POS tagset, we do the contrast experiments with language-specific POS tagsets (English Penn Treebank POS tagset and Chinese Penn Treebank POS tagset) as input in the \textit{ZL} and \textit{LL} models respectively. In the experiments, all settings are the same as introduced above except using different POS tagsets and POS tag embeddings in the network. The results are shown in Table~\ref{tab:tagset}. We can see that both \textit{ZL} and \textit{LL} models with the Universal POS tagset show superior results to the models with different POS tagsets. These results suggest that Universal POS tagsets  provide a good basis for learning cross-lingual knowledge which is important for discourse segmentation.

\section{Related Work}
Recently, much work have been conducted on English EDU segmentation. Most of those segmenters are based on statistical classifiers and leverage many manually extracted features in which lexical syntactic subtree features play an important role~\cite{soricut2003sentence}. Some modifications to the statistic model and features have made great improvements~\cite{fisher2007utility,joty2015codra} and ~\cite{bach2012reranking} achieved the best results ever. Besides, ~\cite{subba2007automatic} first used neural networks on this task and some neural network models have achieved comparable results using less artificial features. Since ~\cite{sporleder2005discourse} see the EDU segmentation task as a sequence labeling problem, we leverage the LSTM-CRF model~\cite{huang2015bidirectional,ma2016end} which is proved effective in this problem.

As for Chinese EDU segmentation, previous work mainly focused on identifying EDU boundaries by punctuations and saw this task as comma classification~\cite{li2012elementary,xue2011chinese,yang2012chinese,xu2013recognizing}. ~\cite{cao2017discourse} claimed to conduct Chinese EDU segmentation based on RST, but in their criteria, no-punctuation EDU boundaries are not considered and only a small corpus is released. Thus, with zero or little segmented Chinese corpus, we attempt to borrow knowledge from abundant English labeled data and design the framework of adversarial neural network to learn the discourse commonality across different languages.

The main motivation of cross-lingual tasks is to use the language commonality between languages or remedy the lack of labeled data in a language.
One kind of methods are to project annotations across parallel corpus~\cite{yarowsky2001inducing,diab2002unsupervised,pado2006optimal,xi2005backoff}. However, it is difficult to obtain parallel corpus. Then, one kind of the alternative methods are statistic model based and require cross-lingual features~\cite{ando2005framework,darwish2013named}; and the other kind of methods are neural network based, since word embeddings in different languages have the capability of representing semantic meanings in the same space. Besides bilingual word embeddings, bilingual character embeddings also improve the performance ~\cite{yang2016multi,cotterell2017low}. However, these work are usually constrained to a language family such as Indo-European languages which share some same characters ~\cite{braud2017cross}. For Chinese and English which belong to different language families, we propose a method to get Universal Chinese POS tags and use adversarial network to solve the language adaption problem inspired by ~\cite{ganin2015unsupervised}, ~\cite{chen2016adversarial} and ~\cite{kim2017cross}.

\section{Conclusions}

In this paper, we propose to segment EDUs in Chinese based on RST and identify those EDU boundaries where there is no punctuations. Further more, we design an adversarial network to exploit abundant English discourse data to help segment Chinese EDUs because of the lack of labeled Chinese data which follows the criterion above. Based on cross-lingual discourse commonality, we use an adversarial language discrimination task to extract common language-independent features and language-specific features which are useful for discourse segmentation. Experimental results verify the efficiency of our models.
\bibliographystyle{aaai}
\bibliography{emnlp2018}

\end{document}